\documentclass{article}

\usepackage[english]{babel}

\usepackage[letterpaper,top=2cm,bottom=2cm,left=3cm,right=3cm,marginparwidth=1.75cm]{geometry}

\usepackage{amsmath}
\usepackage{graphicx}
\usepackage[colorlinks=true, allcolors=blue]{hyperref}
\usepackage{float}
\usepackage{tabularray}

\title{THaLLE: Text Hyperlocally Augmented Large Language Extension - Technical Report}
\author{NLP-Voice Research Lab, KBTG Labs,\\
\textit{KASIKORN Business---Technology Group}\\}

\begin{document}
\maketitle

\begin{abstract}
Recent advancements in Large Language Models (LLMs) have revealed new capabilities and opportunities across the technological landscape. However, the practicality of very large LLMs is challenged by their high compute cost, which does not justify the benefits given their limited capability compared to humans. While smaller, more practical LLMs have shown potential in financial analysis, though they are not yet fully proficient, as evidenced by their near-passing performance on the Chartered Financial Analyst (CFA) exam. In this work, we present Financial Analyst Extension to our Text Hyperlocally Augmented Large Language Extension (THaLLE), a series of 8B LLMs consistently achieving highest performance on mock CFA exams against models of comparable size. We thoroughly document the fine-tuning techniques used to facilitate future research. Additionally, we introduce the use of Flare CFA, a publicly available dataset for evaluating LLMs as a financial advisor.
\end{abstract}

\section{Introduction}

Large Language Models (LLMs) have emerged as leading tools in Natural Language Processing (NLP) due to their exceptional performance across various tasks. The advent of open-source models such as Llama \cite{llama2-touvron2023llama} from Meta, Gemma \cite{team2024gemma} from Google, and Qwen \cite{qwen} from Alibaba has significantly enhanced public access to advanced LLMs. Additionally, low-cost techniques for LLM fine-tuning, such as Low-rank Adaptation (LoRA) \cite{hu2022lora}, have enabled the fine-tuning of these models on consumer-grade hardware, thereby accelerating their development and adoption. LLMs are now utilized in a wide array of applications, ranging from personal assistants, i.e., ChatGPT, to specialized tasks in diverse domains.

In the financial sector, BloombergGPT \cite{wu2023bloomberggpt}, a proprietary LLM trained from the ground up with an infusion of financial data, has demonstrated superior performance on financial benchmarks compared to other models in the market. In the open-source community, the AI4Finance foundation advocates for cost-effective methods, such as fine-tuning foundational models, and has developed an open-source framework for fine-tuning LLMs with financial task-specific data \cite{yang2023fingpt}.

In this paper, we investigate the application of LLMs in financial analysis and advisory roles by evaluating their performance on the Chartered Financial Analyst (CFA) exam, a highly regarded credential for financial advisors. For an LLM, performing on the CFA exam constitutes a multiple-choice reading comprehension (MRC) task within the domain of financial analysis. Our findings reveal that only a few LLMs are capable of passing the CFA exam, primarily due to their generalist training and focus on non-financial tasks. We also provide a comprehensive analysis of fine-tuning strategies aimed at enhancing MRC skills in LLMs, with the goal of improving their performance on finance-specific assessments like the CFA exam.

\section{Background}

In this section, we provide an overview of the CFA exam and review previous studies on the performance of LLMs in this context.

\subsection{The Chartered Financial Analyst Exam}

The CFA program consists of three levels of exams designed to validate the expertise and dedication of finance professionals.
Each level progressively enhances the knowledge and skills acquired in the preceding one, covering a broad spectrum of financial topics.

\subsubsection{Level I Exam}

The Level I exam covers a broad range of fundamental financial concepts and investment tools. It is structured as follows:

\begin{itemize}
    \item Format: 180 multiple choice questions.
    \item Duration: The exam is divided into two sessions of approximately 135 minutes each, with an optional break in between.
    \begin{itemize}
        \item First Session: 2 hours and 15 minutes, comprising 90 multiple-choice questions.
        \item Second Session: 2 hours and 15 minutes, comprising another 90 multiple-choice questions.
    \end{itemize}
    \item Content Area: The exam covers 10 major topics, including ethical and professional standards, quantitative methods, economics, financial reporting and analysis, corporate finance, equity investments, fixed income, derivatives, alternative investments, and portfolio management and wealth planning.
    \item Pass Rate: The 10-year average pass rate for the Level I exam is 41\%, reflecting the challenging nature of the exam and the comprehensive understanding required.
\end{itemize}

\subsubsection{Level II and Level III Exams}

Subsequent levels of the CFA program goes deeper into the analysis and application of investment principles and portfolio management techniques.
The Level II exam focuses on the valuation of various assets and requires candidates to apply their knowledge to real-world scenarios.
The Level III exam emphasizes portfolio management and wealth planning, requiring candidates to synthesize a vast array of financial concepts and apply them to complex situations.

\subsection{Evaluating LLM Finance Proficiency through the CFA Exam}

Previous research \cite{callanan2023can} has explored the potential of LLMs to serve as financial analysts or advisors by evaluating their proficiency using the CFA exam. This study included assessments of ChatGPT and GPT-4 \cite{openai2024gpt4} on Levels I and II of the CFA exam, excluding Level III due to its open-ended questions.

The results indicated that GPT-4 achieved a score exceeding 70\%, which meets the benchmark proposed by experienced CFA examination candidates for certification. This finding suggests that LLMs might possess the capability to pass the CFA exam and become certified financial advisors in the future, thereby expanding their applicability in various financial contexts.

Nonetheless, there remain questions about how open-source LLMs can compete with proprietary commercial-grade LLMs, especially in the context of smaller LLMs.

\section{Experiment}

Our experiment stared with a preliminary evaluation of foundational instruction-tuned generalist models using two CFA exams, as detailed in Section~\ref{sec:exp-setup}. Based on the outcomes of this initial assessment, we selected the best-performing base models for further training. The training data utilized in this phase is described in Section~\ref{sec:training-data}.
We experimented with two fine-tuning routines:
\begin{enumerate}
    \item Supervised Fine-Tuning (SFT)
    \item Direct Preference Optimization (DPO) \cite{NEURIPS2023_a85b405e}
\end{enumerate}
Data augmentation for SFT and DPO are described in Section~\ref{sec:aug-sft}~and~\ref{sec:aug-dpo}, respectively.

% Text Hyperlocally Augmented Large Language Extension

The resulting fine-tuned models, named \textbf{T}ext \textbf{H}yperlocally \textbf{A}ugmented \textbf{L}arge \textbf{L}anguage \textbf{E}xtension (THaLLE), were re-evaluated following the procedures outlined in Section~\ref{sec:exp-setup}. Our ultimate goal is to develop LLMs proficient in multiple subject areas with capabilities in Thai language. In this work, we are specifically investigating the augmentation of the model with financial knowledge. In the future, we intend to further enhance the model's Thai language capabilities. We explored various training configurations and methodologies, some of which were subsequently discarded. These configurations and the rationale behind their rejection are further discussed in the subsections under Section~\ref{sec:results}.

\subsection{Training Data}
\label{sec:training-data}

Our training data consists of 9,429 unique internal CFA exam questions.
Each question includes the question, multiple choices, the correct answer, and the rationale behind each answer.
The dataset covers exams from 2009 to 2019.
Newer exam data were used to build the test set (see Table~\ref{tab:data} and Section~\ref{sec:internal-cfa}).
Human annotators and automated systems checked the exam questions for errors and duplication.

\begin{table}
\centering
\begin{tblr}{
  cells = {c},
  cell{3}{1} = {r=2}{},
  hline{1-3,5} = {-}{},
  hline{4} = {2-4}{},
}
                & \textbf{Years} & \textbf{Sets} & \textbf{Questions} \\
Training data   & 2009-2019      & 80            & 9,426              \\
Evaluation task & 2020           & 8             & 887                \\
                & 2024           & 4             & 360                
\end{tblr}
    \caption{Internal Mock CFA Exam datasets}
    \label{tab:data}
\end{table}

\subsection{Experimental Setup}
\label{sec:exp-setup}

We benchmarked models on two tasks: our internal mock CFA exam and the Flare CFA.
All models are evaluated using the ``Zero Shot'' (ZS) \cite{DBLP:journals/corr/abs-2005-14165} system prompt by \cite{callanan2023can}.
The complete system prompts are detailed in Appendix~\ref{apx:prompt}.
The entire Flare CFA dataset was used as our test set.
See Section~\ref{sec:choice-select} for how we match the output text to one of three answer choices.

For our preliminary evaluation and model selection, we assessed four foundational instruction-tuned models: Llama2-7B\footnote{\href{https://huggingface.co/meta-llama/Llama-2-7b-chat-hf}{meta-llama/Llama-2-7b-chat-hf}} \cite{llama2-touvron2023llama}, Llama3-8B\footnote{\href{https://huggingface.co/meta-llama/Meta-Llama-3-8B-Instruct}{meta-llama/Meta-Llama-3-8B-Instruct}}, Gemma-7B\footnote{\href{https://huggingface.co/google/gemma-7b-it}{google/gemma-7b-it}} \cite{team2024gemma}, Qwen2-7B\footnote{\href{https://huggingface.co/Qwen/Qwen2-7B-Instruct}{Qwen/Qwen2-7B-Instruct}} \cite{qwen}.
The results are presented in Table~\ref{tab:results_zs}.
Based on these results, we selected Llama3-8B Instruct and Qwen2-7B Instruct for further fine-tuning.
We utilize LoRA \cite{hu2022lora} for efficient fine-tuning.

Given the unique nature of fine-tuning on CFA exams in our setup, we also evaluated models that had been continue-pretrained or fine-tuned on other domains to understand the impact of such fine-tuning on performance. We conducted comparisons with:
\begin{enumerate}
    \item LLMs fine-tuned specifically on financial data: FinGPT Multi-Task Llama2-7B\footnote{\href{https://huggingface.co/FinGPT/fingpt-mt_llama2-7b_lora}{FinGPT/fingpt-mt\_llama2-7b\_lora}} \cite{yang2023fingpt}.
    \item Four commercial APIs: GPT-3.5-turbo and GPT-4o from OpenAI, and Gemini-1.5-Flash and Gemini-1.5-Pro from Google.
\end{enumerate}

\subsubsection{Flare CFA}

Flare CFA is a publicly accessible dataset for CFA exam preparation\footnote{\href{https://huggingface.co/datasets/ChanceFocus/flare-cfa}{ChanceFocus/flare-cfa}}.
The data comprises 1,032 questions covering levels I and II of the CFA exam.
This dataset was selected to facilitate comparative analysis in future research.

\subsubsection{Internal Mock CFA Exam Tasks}
\label{sec:internal-cfa}

Our internal mock CFA exam tasks are held-out datasets of our mock exam questions.
The exam tasks contain the newer 887 questions for 2020 and 360 questions for 2024 (see Table~\ref{tab:data}).
Our latest 2024 task is sourced from \href{https://www.cfainstitute.org/en/programs/cfa/prep-providers}{CFA Institute Prep Providers}.

\subsubsection{Choice Selection}
\label{sec:choice-select}

We aim to strike the right balance between accommodating the models and maintaining stringent evaluation standards.
For evaluation purposes, we established a clear criterion that the model must either specify the selected choice (e.g., A, B, C) or recite the content of the choice.

To achieve this, we manually inspected the output patterns of all models and developed a custom matching script to recognize outputs that meet our criteria. For example, given the choice ``A. \$1,308.'' our evaluator will match any of the following responses: ``A'', ``A.'', ``The correct answer is: A'', ``\$1,308.'', etc. If there are multiple matches, we will select the first match.

\subsection{Data Augmentation for SFT}
\label{sec:aug-sft}

In our exploration of fine-tuning LLMs, we identified two key principles that effectively enhanced the effectiveness of our models:

\begin{enumerate}
    \item It is crucial not only to recognize the correct answer but also to understand the underlying reasons for its correctness. This approach ensures a deeper comprehension rather than mere memorization.
    \item Each model possesses unique characteristics, and to achieve optimal results, the training process must align with the model's inherent functioning without causing disruptions.
\end{enumerate}

\subsubsection{Exploration of ``Zero Shot'' Prompts}

The first principle is corroborated by our ``correct answers only'' training data. In this approach, models were trained to predict only the correct answers when provided with a ``Zero Shot'' prompt.
This method, however, resulted in the models failing to acquire new knowledge or generalize effectively, leading to performance degradation compared to the base model.

For Llama3-8B Instruct, we observed that the model naturally attempts to justify its choices after selecting an answer. Adhering to the model's intrinsic behavior (as per the second principle), we found that incorporating reasoning in the manner the model would naturally output (i.e., adding reasons afterward) enhanced its learning and performance.

Conversely, Qwen2-7B Instruct did not exhibit the same behavior of adding reasons when given a ``Zero Shot'' prompt. This model achieved better performance when trained with the modified ``Zero Shot'' prompt that requested the model to specify reasons after answering, similar to the approach used for Llama3. Training on data formatted this way resulted in improved performance.

\subsubsection{Incorporating ``Chain-of-Thought'' Prompts for Variety}

In addition to the training samples with ``Zero Shot'' prompts and data that already contain both the chosen answer and the reasoning, we experimented with adding ``Chain-of-Thought'' \cite{DBLP:journals/corr/abs-2201-11903} prompts.
Unlike ``Zero Shot'' prompts, ``Chain-of-Thought'' prompts require the model to produce the reasoning before providing an answer.
We found that this approach also enhanced the model's performance.

\subsection{Self-Supervised Data Augmentation for DPO}
\label{sec:aug-dpo}

Building on the first principle discussed in the previous Section~\ref{sec:aug-sft}, it is essential to identify and understand the faulty reasoning that leads to incorrect answers.
To enhance the efficacy of Direct Preference Optimization (DPO), we first allowed the model to answer training questions and collect incorrect responses.
Similar to data augmentation for SFT, the DPO also include both ``Zero Shot'' prompts and ``Chain-of-Thought'' prompts in training samples for each exam question.

\subsubsection{Samples with ``Zero Shot'' prompts}

This initial phase utilized a ``Zero Shot'' prompt to quickly identify and discard correct answers.
Notably, Llama3 inherently tends to produce reasoning without additional prompting, whereas Qwen2 requires a modified prompt instructing it to add reasoning.

To increase the yield of incorrect answers, we performed inference with sampling and a temperature setting of 1.0. Each question was subjected to multiple passes to gather at least one incorrect reasoning for each incorrect answer, incrementing the temperature by 0.1 with each pass up to a maximum of 1.2. For any incorrect choices not produced by this method, we manually forced the model to predict the incorrect output and generate a supporting reason. This approach ensured that for each question, we obtained two incorrect reasonings, since in CFA exams, there are always three answer choices, one of which is correct.

These incorrect responses constituted two training samples using the ``Zero Shot'' prompt. In these samples, the ``chosen output'' was the correct choice followed by the correct reasoning, while the ``rejected output'' was one of the incorrect choices followed by the corresponding incorrect reasoning.

\subsubsection{Samples with ``Chain-of-Thought'' prompts}

For the ``Chain-of-Thought'' prompt, we employed a similar process to obtain incorrect reasonings for incorrect answers. However, for those incorrect answers not produced by this method, we faced the non-trivial challenge of forcing the model to generate a specific incorrect answer. To address this, we borrowed incorrect reasonings from the ``Zero Shot'' prompt phase.

In this phase, each training sample consisted of a ``Chain-of-Thought'' prompt. In these samples, the ``chosen output'' included the correct reasoning followed by the correct choice, while the ``rejected output'' included one of the previously identified incorrect reasonings followed by the incorrect choice. 

This method of self-supervised data augmentation allowed us to systematically improve the model's understanding and reasoning capabilities, thereby enhancing its overall performance through DPO.

\section{Results}
\label{sec:results}

\subsection{Zero Shot Performance}

As shown in Table~\ref{tab:results_zs}, the latest instruction-following models (Gemma, Llama3, and Qwen2, released in 2024) excel in the CFA exam for both the Flare CFA and our internal mock exam.
These findings highlight the significant advancements within the open-source LLM community, with some newer smaller models even surpassing the performance of proprietary models like GPT-3.5-turbo from OpenAI.

Despite our efforts to accommodate FinGPT by employing custom prompts and choice selection, this model was unable to meet our evaluation criteria with consistently (see Section~\ref{sec:choice-select}).
Consequently, FinGPT results are omitted from Table~\ref{tab:results_zs}.
Example output from FinGPT is shown in Appendix~\ref{apx:outputs}.

The general guideline for the CFA exam’s minimum passing score is typically around 70\%. The specific estimations for the minimum passing scores for 2020 and 2024 are approximately 65\%\footnote{\href{https://300hours.com/cfa-passing-score/}{300Hours - CFA Passing Score: Here’s the Latest MPS Estimates}}.
When evaluating the models against these scoring criteria, both GPT-4o and THaLLE achieved passing scores in all three mock exams: Flare CFA, Internal 2020 CFA, and Internal 2024 CFA.
The Gemini-1.5-Flash and Gemini-1.5-Pro successfully passed the Flare CFA mock exam, and the Gemini-1.5-Pro also met the passing score for the Internal 2020 CFA mock exam.

\begin{table}[ht]
    \centering
    \begin{tblr}{
      cell{1}{1} = {r=2}{},
      cell{1}{2} = {c=2}{c},
      cell{1}{4} = {r=2}{c},
      cell{3}{1} = {c=3}{},
      cell{8}{1} = {c=3}{},
      column{2} = {r},
      column{3} = {r},
      column{4} = {r},
      hline{1,3,8,13,17,18,22} = {-}{},
    }
    \textbf{Model}                                     & \textbf{Internal Mock CFA} &          & \textbf{Flare CFA} \\
                                                       & \textbf{2020}         & \textbf{2024} &                    \\
    \textbf{Commercial APIs}                           &                       &               &                    \\
    \texttt{gpt-3.5-turbo-0125}~                       & 0.5458                & 0.5027        & 0.6366             \\
    \texttt{gemini-1.5-flash-001}                      & 0.6271                & 0.6278        & 0.7355             \\
    \texttt{gemini-1.5-pro-001}                        & 0.6780                & 0.6444        & 0.7829             \\
    \texttt{gpt-4o-2024-05-13}                         & \textbf{0.8000}       & \textbf{0.8055} & \textbf{0.8789}  \\
    \textbf{Open instruction-tuned foundational LLMs}  &                       &               &                    \\
    Llama-2-7B-chat                                    & 0.3774                & 0.3639        & 0.4264             \\
    Gemma-7B-instruct                                  & 0.5107                & 0.5333        & 0.6027             \\
    Llama3-8B-instruct                                 & 0.5424                & 0.5222        & 0.6386             \\
    Qwen2-7B-instruct                                  & \textbf{0.5740}       & \textbf{0.5583} & \textbf{0.6831}  \\
    THaLLE-SFT (Llama3-8B-instruct)~~~~~                & 0.5751                & 0.5500        & 0.6570            \\
    THaLLE-DPO (Llama3-8B-instruct)                     & 0.5582                & 0.5306        & 0.6492            \\
    THaLLE-SFT (Qwen2-7B-instruct)                      & \textbf{0.6678}       & \textbf{0.6500} & \textbf{0.7171} \\
    THaLLE-DPO (Qwen2-7B-instruct)                      & 0.5887                & 0.5833        & 0.6870            \\                
    \end{tblr}
    \caption{Models performance on CFA Exams with Zero Shot system prompt.}
    \label{tab:results_zs}
\end{table}

\subsection{Supervised Fine-tuning (SFT)}

We conducted Supervised Fine-tuning experiments on instruction-following models for MRC tasks, with our internal mock CFA exams.
The results show improvement in both task-following aspects and achieve higher scores across the test set, as shown in Table~\ref{tab:results_zs}.

\subsubsection{Over-fitting on SFT}

From our experiments, we found that the model typically converges within 1-3 epochs.
Prolonged training can lead to overfitting, which is detectable through validation set evaluations.
This tendency is especially pronounced when fine-tuning Llama3-8B instruct, likely due to the model's broad range and extensive pretraining.
To mitigate overfitting, we recommend incorporating a validation set and an early stopping mechanism in the training pipeline.

\subsubsection{Effect of Prompt Loss Masking}

Prompt loss masking is a technique employed to exclude the loss associated with the model's next-token prediction for the prompt segment. This approach allows the training of LLMs to concentrate on the prediction of the target text portion that is essential for the model's learning process. 

Our investigation indicates that applying prompt loss masking during SFT results in a notable enhancement in overall performance. Consequently, we advocate for the evaluation of prompt loss masking in MRC tasks when fine-tuning LLMs.

\subsection{Direct Preference Optimization (DPO)}

We conducted Direct Preference Optimization experiments on instruction-following models for MRC tasks, with our internal mock CFA exams. The results show improvement in both task-following aspects and achieve higher scores across the test set, as shown in Table~\ref{tab:results_zs}.

\subsubsection{Overfitting and DPO}

Models trained using DPO exhibit a slower rate of overfitting compared to those trained with SFT. We hypothesize that this difference arises from the regularization effect provided by the reference model in DPO. This hypothesis is supported by our observations during the fine-tuning of the Llama3-8B model. In our initial experiments, the SFT approach led to overfitting, resulting in performance degradation relative to the original model when the configuration was not properly adjusted. Conversely, the DPO method consistently achieved a performance improvement without the same overfitting issues.

\subsubsection{Base Model}

Our observations reveal that fine-tuning performance using DPO across various models requires substantial adjustments in hyperparameters.
In our experiments, DPO techniques applied to the Llama3-instruct model without SFT were effective, whereas they exhibited limited efficacy with the Qwen2-instruct model.
We hypothesize that the differing training procedures utilized for the foundational and instruct models significantly influence the success of DPO.

\subsection{Training and Optimization Techniques}

\subsubsection{Effect of LoRA Target Modules}

As suggested in the \cite{hu2022lora}, increasing the number of target modules in LoRA generally leads to improved fine-tuning performance.
Our investigation support this claim; utilizing all-linear target modules for LoRA weights enabled faster convergence during training at any LoRA rank.
Conversely, experiments with LoRA employing only basic \texttt{q\_proj} and \texttt{v\_proj} target modules yielded suboptimal performance.
Therefore, we recommend the use of all-linear target modules for MRC task training.

\subsubsection{LoRA Rank}

We examined the impact of LoRA rank on fine-tuning performance. Our findings indicate a positive correlation between LoRA rank and performance up to a rank value of 32. Increasing the rank value beyond 32, however, does not enhance performance. We speculate that for tasks involving moderate English language complexity, such as our CFA exam tasks, a LoRA rank of 32 is optimal. Consequently, we suggest that implementers experimenting with similar tasks and data begin their hyperparameter optimization with a LoRA rank of 32.

\subsubsection{Grouping by Length}

While grouping data by length improves the efficiency of training, it reduces the randomness of the data order, which in turn diminishes model performance. This effect also complicates the interpretation of training statistical graphs. We suggest using \textit{group\_by\_length=False} during the early development phase, which requires extensive hyperparameter tuning.

\subsubsection{System Prompt Training}

We investigate the effect of removing system prompts from training prompts on predictive performance. Our results indicate that the model performs consistently worse without the system prompt included in training. We recommend always incorporating system prompts in training samples when fine-tuning with both SFT and DPO.

\subsubsection{Reasoning in MRC}

As previously discussed in Sections~\ref{sec:aug-sft} and \ref{sec:aug-dpo}, incorporating reasoning is crucial for our CFA exam. Therefore, we conclude that it is highly beneficial for LLM fine-tuning on MRC.
Our experiments demonstrate that including Chain-of-Thought prompting in the training dataset consistently outperforms models trained solely under ``Zero Shot'' settings, where only the answer is predicted.
We even achieved success with a modified ``Zero Shot'' prompt, where the model is instructed to provide an answer followed by reasoning to support it.
We recommend the addition of Chain-of-Thought data or, at the very least, incorporating reasoning into training prompts to improve overall performance.

\subsubsection{Few-Shot Training}

Despite success in few-shot inference scenarios across various tasks, training a sequence of multiple questions and answers at once results in suboptimal outcomes. We suggest avoiding the training of multiple questions in the few-shot manner for LLMs in both SFT and DPO.

\section{Conclusion}
\label{sec:conclusion}

In this work, we conducted thorough experiments on supervised fine-tuning and direct preference optimization techniques for instruction-following LLMs in MRC settings, such as the CFA exam.
Our research assessed various training pipeline variants, reported their effects, and provided mitigation strategies and suggestions.

A significant milestone was the development of THaLLE, the first model from a Thai company to pass the CFA exam, opening opportunities for open-source LLM finance advisors/analysts. Fine-tuning general open-source LLMs demonstrated notable advantages, including superior performance and significantly lower costs compared to proprietary models like BloombergGPT. This efficiency allows for rapid fine-tuning, crucial for the dynamic nature of financial data.

Our findings support the idea that LLMs can be effectively developed for financial applications \cite{yang2023fingpt}, despite not finding immediate success with their pre-trained model, FinGPT. However, while using the CFA exam as an evaluation proxy is motivated by its real-world application, we cannot conclude that LLMs are ready to function as working finance advisors. This requires more in-depth and detailed assessments in future work.

Our research offers valuable insights into fine-tuning instruction-following LLMs and highlights the potential for cost-effective, high-performance LLM development in the finance industry.
We hope this work aids researchers and developers in avoiding common pitfalls and accelerates open-source LLM development.

\section{Future Work}

We plan on applying DPO to the SFT model.
Previous research indicates that DPO on non-SFT models may yield suboptimal results \cite{NEURIPS2023_a85b405e}.
In addition, we plan to investigate the efficacy of various prompt types, such as Chain-of-Thought and few-shot prompts.

As outlined in Section~\ref{sec:conclusion}, evaluating LLMs using mock CFA exams does not conclusively guarantee their optimal performance in real-world applications.
Therefore, our future research will involve testing LLMs in more practical scenarios and conducting evaluations in real-world settings.
Our ultimate objective is to develop LLMs that are proficient in multiple subject areas and capable of utilizing the Thai language effectively.
To this end, we will perform comparative evaluations to assess the impact of downstream training, explore methods to mitigate any adverse effects, and optimize trade-offs.
Additionally, we will investigate novel techniques for weight merging to efficiently integrate diverse capabilities.

\section*{Contributions and Acknowledgments}

We extend our gratitude to the executive team for their leadership and support, and to the data team for their efforts in cleaning and validating the internal datasets.\newline
\newline
\textit{Core Contributors:} Danupat Khamnuansin, Atthakorn Petchsod, Anuruth Lertpiya, Pornchanan Balee, Thanawat Lodkaew
\newline
\textit{Data Team:} Panumate Chetprayoon,  Tharathip Chuenprasaeng, Thanawin Preechapate, Pattanun Kamsat, and anonymous annotators
\newline
\textit{Project Managers:} Patcharin Areewong
\newline
\textit{Executive Leadership\footnote{\label{footnote:name}sort by alphabetical order}:} Monchai Lertsutthiwong, Tawunrat Chalothorn, Thadpong Pongthawornkamol

\newpage
\bibliographystyle{unsrt}
\bibliography{ref}

\newpage

\section*{Appendix}
\appendix

\section{System Prompt}
\label{apx:prompt}

\begin{table}[H]
    \centering
    \begin{tabular}{|p{3.25cm}|p{11.5cm}|}
        \hline
        \textbf{Type} & \textbf{System Prompt} \\
        \hline
        Zero Shot \cite{callanan2023can} & {``You are a CFA (chartered financial analyst) taking a test to evaluate your knowledge of finance. You will be given a question along with three possible answers (A, B, and C).\symbol{92}nIndicate the correct answer (A, B, or C).''} \\
        \hline
        modified Zero Shot \newline (fine-tuning Qwen2) & {``You are a CFA (chartered financial analyst) taking a test to evaluate your knowledge of finance. You will be given a question along with three possible answers (A, B, and C).\symbol{92}nIndicate the correct answer (A, B, or C).\symbol{92}nAlso provide the reason to support your answer.''} \\
        \hline
        Chain-of-Thought & {``You are a CFA (chartered financial analyst) taking a test to evaluate your knowledge of finance. You will be given a question along with three possible answers (A, B, and C).\symbol{92}nBegin your answer with ``Thought: '', think as much as you needed, then finalize your answer using ``Therefore, the correct answer is: <FINAL\_ANSWER>''.''} \\
        \hline
    \end{tabular}
    \caption{Prompts used in our experiments}
    \label{tab:prompt}
\end{table}

\section{Model Output Examples}
\label{apx:outputs}

Table~\ref{tab:outputs} shows examples model output when given ``Zero Shot'' prompt on a mock CFA exam question.

System Prompt: ``You are a CFA (chartered financial analyst) taking a test to evaluate your knowledge of finance. You will be given a question along with three possible answers (A, B, and C).\symbol{92}n\symbol{92}nIndicate the correct answer (A, B, or C).''

User Prompt: ``Question:\symbol{92}nWhich of the following return calculating methods is best for evaluating the annualized returns of a buy-and-hold strategy of an investor who has made annual deposits to an account for each of the last five years?\symbol{92}nA. Geometric mean return.\symbol{92}nB. Arithmetic mean return.\symbol{92}nC. Money-weighted return.''

\begin{table}[H]
    \centering
    \begin{tabular}{|p{3.75cm}|p{11cm}|}
        \hline
        \textbf{Model} & \textbf{Output} \\
        \hline
        \texttt{gpt-3.5-turbo-0125}    \newline Zero Shot & {``A. Geometric mean return.''} \\
        \hline
        \texttt{gemini-1.5-flash-001}  \newline Zero Shot & {``The correct answer is **C. Money-weighted return**. Here's why:\symbol{92}n\symbol{92}n* **Money-weighted return** considers the timing and amount of cash flows into and out of the investment. This is crucial for a buy-and-hold strategy with annual deposits, as the timing of those deposits significantly impacts the overall return.\symbol{92}n\symbol{92}nLet's look at why the other options are not the best fit:\symbol{92}n\symbol{92}n* **Geometric mean return** is best for calculating the average return over multiple periods when the investment is held for the entire period. It doesn't account for cash flows during the holding period.\symbol{92}n* **Arithmetic mean return** is a simple average of returns over multiple periods. It doesn't consider the timing or size of cash flows, making it unsuitable for evaluating a strategy with regular deposits.\symbol{92}n\symbol{92}n**In summary:**  The money-weighted return is the most appropriate method for evaluating the annualized returns of a buy-and-hold strategy with annual deposits because it accurately reflects the impact of those deposits on the overall return. \symbol{92}n''} \\
        \hline
        \texttt{gemini-1.5-pro-001}    \newline Zero Shot & {``The correct answer is **(C) Money-weighted return.**\symbol{92}n\symbol{92}nHere's why:\symbol{92}n\symbol{92}n* **Money-weighted return (MWR)** calculates the internal rate of return (IRR) of an investment, considering all cash flows. In a buy-and-hold strategy with regular deposits, the MWR accurately reflects the timing and impact of each deposit on the overall return.\symbol{92}n\symbol{92}n* **Geometric mean return** is useful for calculating the average return over multiple periods, assuming the returns are compounded. However, it doesn't account for the varying amounts invested at different times, making it less suitable for this scenario.\symbol{92}n\symbol{92}n* **Arithmetic mean return** is a simple average of returns and doesn't consider compounding or the timing of cash flows, making it inappropriate for evaluating a strategy with regular deposits. \symbol{92}n\symbol{92}n**In summary:** When evaluating a buy-and-hold strategy with periodic investments, the money-weighted return provides the most accurate representation of the investor's actual returns by considering the timing and size of all cash flows. \symbol{92}n''} \\
        \hline
        \texttt{gpt-4o-2024-05-13}     \newline Zero Shot & {``A. Geometric mean return.''} \\
        \hline
        Qwen2-7B-instruct \newline Zero Shot & {``A''} \\
        \hline
        FinGPT (Llama2-7B)\newline Zero Shot & Model outputs {\texttt{</s>}}, which is equivalent to empty string. \\
        \hline
        THaLLE-SFT (Qwen2) \newline Zero Shot & {``A. Geometric mean return.''} \\
        \hline
        THaLLE-SFT (Qwen2) \newline modifed Zero Shot & {``A. Geometric mean return.\symbol{92}n\symbol{92}nThe geometric mean return is the appropriate method to calculate the annualized return of a series of investments that are held for different lengths of time. The money-weighted return is the appropriate method when cash flows are irregular. The arithmetic mean return is the simple average of all the returns and does not take into account the compounding effect.''} \\
        \hline
        THaLLE-SFT (Qwen2) \newline Chain-of-Thought & {``Thought: The geometric mean return is the appropriate method for computing the average return of a series of periodic returns. The arithmetic mean return would be incorrect because it does not take into account the compounding effect of the returns. The money-weighted return is calculated from the cash flows, not from the periodic returns.\symbol{92}n\symbol{92}nTherefore, the correct answer is: A. Geometric mean return.''} \\
        \hline
    \end{tabular}
    \caption{Examples of model output when prompted to answer CFA exam questions.}
    \label{tab:outputs}
\end{table}

\end{document}